\pdfoutput=1
\documentclass[10pt]{article}

\usepackage[a4paper,textwidth=18cm,textheight=24cm,top=2.85cm, bottom=2.85cm, left=1.5cm, right=1.5cm]{geometry}

\usepackage{icdp2009}

\makeatletter
\long\def\@makecaption#1#2{%
\vskip\abovecaptionskip
\sbox\@tempboxa{#1. #2}%
\ifdim \wd\@tempboxa >\hsize
#1. #2\par
\else
\global \@minipagefalse
\hb@xt@\hsize{\box\@tempboxa\hfil}%
\fi
\vskip\belowcaptionskip}
\makeatother

\usepackage{lmodern}
\usepackage[multiple]{footmisc}
\usepackage{graphicx}
\usepackage{times}
\usepackage[numbers,sort&compress]{natbib}
\renewcommand{\thefootnote}{\fnsymbol{footnote}}
\begin{document}
\noindent

\title{Automated detection of smuggled high-risk security threats using Deep Learning}
\authorname{N. Jaccard$^{1}$, T.W. Rogers$^{1,2}$, E.J. Morton$^{3}$, L.D. Griffin$^{1}$\footnotemark[1]}
\authoraddr{1. Department of Computer Science, University College London, UK}
\secondauthoraddr{2. Department of Security and Crime Sciences, University College London, UK}
\thirdauthoraddr{3. Rapiscan Systems Ltd., Stroke-on-Trent, UK}
\maketitle

\footnotetext[1]{Correspondence: l.griffin@cs.ucl.ac.uk}
\footnotetext[2]{{Please note, we use the term ``small metallic threats'' as we do not wish to make our research results easily discoverable by malicious actors through keyword searching. However, the threats in question are similar in form to hand drills}}

\abstract
The security infrastructure is ill-equipped to detect and deter the smuggling of non-explosive devices that enable terror attacks such as those recently perpetrated in western Europe. The detection of so-called ``small metallic threats'' (SMTs) in cargo containers currently relies on statistical risk analysis, intelligence reports, and visual inspection of X-ray images by security officers. The latter is very slow and unreliable due to the difficulty of the task: objects potentially spanning less than 50 pixels have to be detected in images containing more than 2 million pixels against very complex and cluttered backgrounds. In this contribution, we demonstrate for the first time the use of Convolutional Neural Networks (CNNs), a type of Deep Learning, to automate the detection of SMTs in fullsize X-ray images of cargo containers. Novel approaches for dataset augmentation allowed to train CNNs from-scratch despite the scarcity of data available. We report fewer than 6\% false alarms when detecting 90\% SMTs synthetically concealed in stream-of-commerce images, which corresponds to an improvement of over an order of magnitude over conventional approaches such as Bag-of-Words (BoWs). The proposed scheme offers potentially super-human performance for a fraction of the time it would take for a security officers to carry out visual inspection (processing time is approximately 3.5s per container image).\let\thefootnote\relax\footnotetext{This paper is a preprint of a paper submitted to the Imaging for Crime Detection and Prevention conference and is subject to Institution of Engineering and Technology Copyright. If accepted, the copy of record will be available at IET Digital Library}

\section{Introduction}
At the turn of the 21st century, the \emph{modus operandi} of terrorist attacks in the West such as those in Madrid and London, often relied on the use of explosives. However, with the 2008 Mumbai attacks, Western governments have become increasingly concerned about the possibility of ``Mumbai-style'' attacks. These concerns have been further compounded by the recent events in Tunisia, France, and Belgium. These attacks have shown the devastation possible using only so-called ``Small Metallic Threats'' (SMTs)\footnotemark[2]. It is thus necessary to detect and disrupt the SMT smuggling routes to prevent such devices from getting into the hands of would-be terrorists. While airport and aviation security is almost total, other routes such as road, rail, and maritime remains vulnerable to smuggling attempts. Automated detection of such threats remains an open research endeavor.

Potentially, any one of the hundreds of millions of cargo containers shipped globally each year could be exploited by malicious actors to smuggle security threats, such as SMTs, across borders. Currently, statistical risk analysis and intelligence reports drive targeted inspection efforts~\cite{King2005,Weele2010} but those measures are unlikely to remain sufficient against increasingly sophisticated smuggling schemes. Instead, security agencies are pushing for a significant step-up in non-invasive inspection capabilities~\cite{archick2010us}, with transmission X-ray scanners being the most commonly used imaging modality for cargo containers~\cite{McDaniel2005a}. However, current detection capabilities are not adequate to accommodate the increasing volumes of images. Indeed, the manual inspection of X-ray security imagery is a painstaking process~\cite{Wolfe2013}. Images of cargo containers pose the most difficult inspection challenge: threats (e.g. SMTs) are often very small relative to the image size (e.g. 0.1\% of pixels in a $2600{\times}850$ pixel image is typical); threats concealed within legitimate cargo can be almost undetectable to the naked eye due to complex or dense obscuration; and the diversity of objects that can be found in a container make it impossible for the officers to learn the complete range of appearances for benign items.

In order to alleviate these issues, we propose the use of computer vision and machine learning techniques for the automated detection of SMTs in single-energy single-view X-ray cargo images. This approach provides multiple advantages over manual inspection: i) orders of magnitude reductions in inspection times; ii) improved and potentially super-human detection performance; iii) computing power can be scaled up to meet the increasing volumes of images to inspect; and iv) greatly simplifies scanning logistics by offering consistent processing times. However, most state-of-the-art computer vision methods were developed for natural imagery (photography) first and foremost, from which X-ray images differ significantly due to their translucency, noise levels, clutter, and skewed perspective~\cite{Zhang2014,Ba2011,McDaniel2005a}. 

Conventional computer vision methods that rely on ``hand-crafted'' features designed for natural images are thus unlikely to perform optimally when applied to X-ray images. Rather than adapting existing features, or deriving novel ones, one can instead use representation-learning methods whereby features that optimize the separation of different image classes are learnt directly from training images. Convolutional Neural Networks (CNNs), part of a family of learning algorithms known as Deep Learning (DL), are representation-learning methods~\cite{LeCun2015} that were recently shown to significantly outperform other computer vision approaches~\cite{He2015}. The main barrier to the application of CNNs to X-ray imagery is the scarcity of training images: threats are rare in Stream-of-Commerce (SoC) and acquiring images of staged smuggling attempts is prohibitively costly and time-consuming. In other fields, this issue was addressed by augmenting the training dataset through the use of synthetic examples~\cite{2014arXiv1406.2227J,2016arXiv160301312L}. In this contribution, we employ a dataset augmentation method where physically-accurate images are synthesised by projection of threats into SoC images~\cite{rogers2016threat}, enabling the generation of very large number of \emph{de-novo} examples with very diverse appearance. We also show that log-transforming input X-ray images significantly improves SMT detection performance.

This paper is structured as follows. First, related research is discussed in Section~\ref{relatedWork}. The methods used, including data set augmentation, CNN architectures, and performance evaluation, are described in Section~\ref{method}. Our main findings are presented and discussed in Section~\ref{results} before concluding in Section~\ref{conclusion}.
\section{Related work}
\label{relatedWork}
The urgent need for robust methods to fill the detection capability gap is not being matched by the current research output in automated analysis for X-ray cargo images, which was recently throughly documented and reviewed in Ref.~\cite{rogers2016automated}. Impressive performance has been reported for the detection of security threats (including SMTs)~\cite{Turcsany2013,Ba2011,Flitton2015,akccaytransfer} in baggage X-ray images, partly made possible by the small dimensions and complexity (e.g. constrained packing and low diversity of objects) of bags, as well as the availability of data-rich and high resolution imaging modalities, including multi-view and volumetric scanning. In comparison, scenes in cargo container imagery tend to be much larger and complex, little constraints on how goods are arranged, and a very large and diverse space of possible objects (i.e. any object that physically fits into a cargo container). As such, it is expected that performance for cargo images would be in general lower than what has been reported for baggage imagery.

Two methods for the automated verification of manifest information based on machine vision algorithms were described~\cite{Zhang2014,Tuszynski2013b}. Zhang and colleagues~\cite{Zhang2014} developed an approach for the classification of X-ray cargo images into 22 categories (e.g. grain, tires) based on a Bag-of-Words learnt from responses to Leung-Malik filters. The categories of 51\% and 78\% of images were in the top and top three categories predicted by their scheme, respectively. Tuszynski \emph{al.}~\cite{Tuszynski2013b} computed a city block distance to measure the distance between intensity histograms of log-transformed images and those of training images for each of the 92 categories considered. Based on this distance, the scheme proposed by the author was able to verify that a given image was associated with the correct category with 48\% accuracy and a 5\% false alarm rate, which was a significant improvement over chance. When using the same approach to predict the category of the imaged container, the category of 31\% of the imaged container was correctly predicted, and it was in the top five predictions 65\% of the time.

Approaches were also proposed for empty container verification, which is useful to avoid unnecessary subsequent processing and to detect ``false empties''~\cite{rogers2015detection,Andrews2016}. Rogers \emph{et al.}~\cite{rogers2015detection} classified cargo container images as empty or non-empty based on a set of fixed geometric features (oriented Basic Image Features), image moments, and the coordinates of sampled windows learnt by a Random Forest classifier. The use of windows coordinates as a feature encouraged the classifier to learn location-dependent ranges of appearance. The authors reported 99.3\% detection with 0.7\% false alarms on SoC images, and 90\% detection with 0.5\% false alarms for synthetic adversarial examples where objects equivalent to 1L of water were placed in empty containers. Andrews \emph{et al.}~\cite{Andrews2016} used anomaly detection techniques, based on features extracted from the hidden layer of an auto-encoder, to perform the same task, achieving 99.2\% accuracy by training the system solely on down-scaled images of empty containers and considering non-empty images as anomalies. 

We recently reported on the first use of Deep Learning for the detection of cars in complex X-ray imagery and reported that Convolutional Neural Networks (CNNs) significantly outperformed conventional Bag-Of-Words (BoW) methods with a 100\% detection rate and fewer than 1-in-454 false alarms raised from containers without a car present~\cite{2016arXiv160608078J}. The scheme correctly detected cars in cases where they were almost completely occluded by other goods. ``Small Metallic Threats'' (SMTs) are significantly more challenging to detect than cars: i) small form factors, ii) very large number of models and manufacturers, iii) appearance close to that of legitimate cargo, and iv) unrestricted orientation. We previously presented preliminary results for the detection of SMTs in small ${256{\times}256}$ patches at a conference, with the additional caveat that the most challenging cases (dense backgrounds) were left-out of the analysis~\cite{Jaccard2016}. In this contribution, we present results for the automated detection of SMTs in \emph{full-size} images and with performance evaluated across all types of background. In addition, we explore various network architectures and compare performance between pre-trained and trained-from-scratch CNNs.

\section{Methods}
\label{method}
\subsection{Dataset and Data Augmentation}
Benign images used for this work were acquired using a Rapiscan Eagle\textregistered R60 rail scanner equipped with a 6MV linac source. Images are 16-bit, grayscale, and their size varies between $1290{\times}850$ and $2570{\times}850$ pixel for 20 and 40ft long cargo containers, respectively. The resolution is ${\approx}6$mm pixel$^{-1}$ in the horizontal direction. The images were randomly sampled from Stream-of-Commerce (SoC) images acquired over several weeks and can be empty (${\approx}20\%$ of the dataset) or contain pallets of commercial cargo, heavy machinery and industrial equipment, household goods, and bulk materials.

SMT images were acquired separately and are part of a proprietary dataset. In total approximately 700 instances of SMTs were available across all types, models, and poses. The original scans were not used directly, but instead individual instances were extracted to create a database of SMTs, which in turn was used to synthesise \emph{de-novo} examples for training. The synthesis process, based on the multiplicative nature of X-ray transmission image formation, was described elsewhere~\cite{rogers2015detection,Jaccard2016} and has recently been shown to be indistinguishable from real threat imagery~\cite{rogers2016threat}. In short, a patch containing a single SMT instance was first cropped out of the full-size image. Pixel-wise segmentation of SMT instances was carried out manually, resulting in a SMT binary mask. Background correction was performed by dividing the cropped patch by the mean intensity of pixels outside of the SMT binary mask. If unrelated objects or structures appeared in the patch (e.g. parts of other SMTs or supporting structures), the corresponding pixels were also ignored during background correction. The SMT instance can then be projected into another X-ray image by intensity multiplication. 

Projecting the same SMT instance into different images results in vastly different appearances due to the translucency property of X-ray images. The dataset is made more diverse by the injection of realistic variations such as intensity scaling and flipping.

In order to train the classification scheme, $1{\times}10^5$ SoC images were randomly sampled and SMT instances were projected into half of them. 75\% and 25\% of the dataset was used for training and testing, respectively. There was no overlap between training and testing data, neither in the SoC backgrounds used, nor in the SMT instances projected.

\subsection{Performance evaluation}
For performance evaluation, it was assumed that images of the negative class (i.e. images without SMTs) would generally produce lower image scores $p_{I}$ than images of the positive class (i.e. images containing \emph{at least} one SMT). Various performance metrics were computed based on $p_{I}$ scores obtained for images in the test set, including the area under the ROC curve (AUC) and the H-measure. The latter is a variant of the AUC that addresses issues related to underlying cost functions~\cite{hand2009measuring,Hand2012}. In addition to the AUC and H-measure, the false positive rate (FPR) was determined by thresholding $p_{I}$ using the $t_{90}$ threshold that resulted in a 90\% detection rate.

\subsection{Classification scheme}
The detection of SMTs in X-ray cargo images was implemented as a binary classification task, with benign images (no SMTs) taken as the negative class and SMT images (at least one SMT) taken as the positive class. The image classification scheme is window-based: i) small windows are densely sampled with a stride $s$; ii) windows are classified and given a score $p_{w,i}$ (the confidence that the $i$-th window contains a SMT or part thereof); iii) whole-image score $p_{I}$ is computed as the maximum score across all windows; iv) image class prediction is obtained by comparing $p_{I}$ with a threshold $t_{90}$. Training was thus conducted on a per-window basis, while performance evaluation was carried out based on full-size scanner images.
\begin{figure}[h]
\centering
\includegraphics[width=0.98\linewidth]{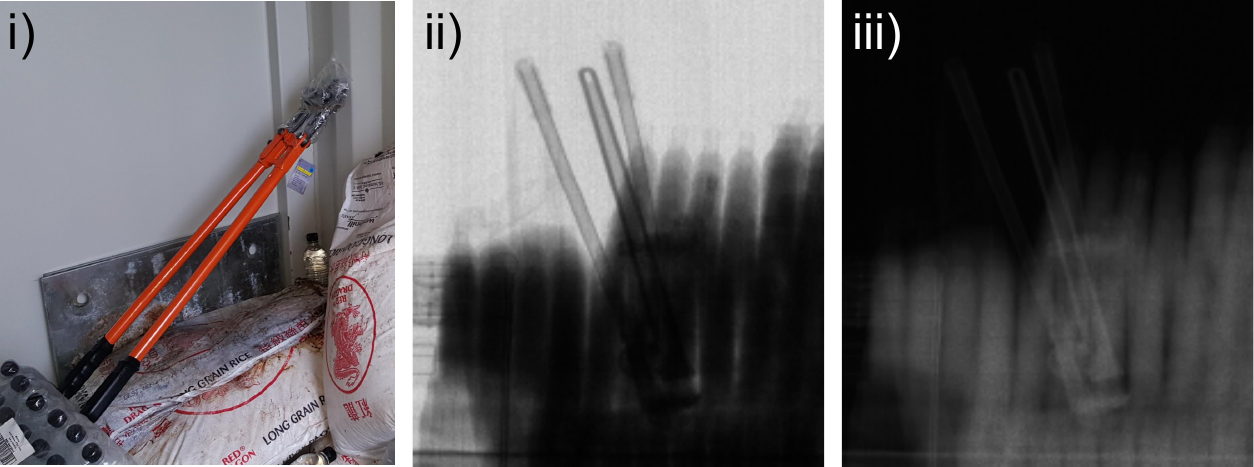}
\caption{Effect of the log-transform on X-ray images of bolt cutters. Image i) shows a photograph of the imaged bolt cutters, while ii) and iii) show the raw intensity and log-transformed images, respectively. Note: bolt cutters are used for illustration, the SMTs of interest are often much smaller.}
\label{fig:BoltCutters}
\end{figure}

For classification by CNNs, the window size was $256{\times}256$ pixels and the stride $s$ was 64 pixels. When comparing with Bag-of-Words (BoW) approaches, the window size was reduced to $64{\times}64$ pixels and the stride $s$ to 32 pixels to maximize BoW performance. 

Prior to classification, images were preprocessed~\cite{rogers2015detection,Jaccard2016}: i) black columns produced by faulty detectors or source misfires were removed, ii) source intensity variations were corrected by normalization based on air intensity values, and iii) salt-and-pepper pixels were replaced by the local median intensity. Raw intensity experiments use preprocessed images as input. When specified, images were log-transformed prior to classification; this transform is frequently used to facilitate detection of concealed items by security officers during visual inspection (Fig~\ref{fig:BoltCutters}) and was also previously applied to automated classification~\cite{Tuszynski2013b}.
\begin{figure}
\centering
\includegraphics[width=0.94\linewidth]{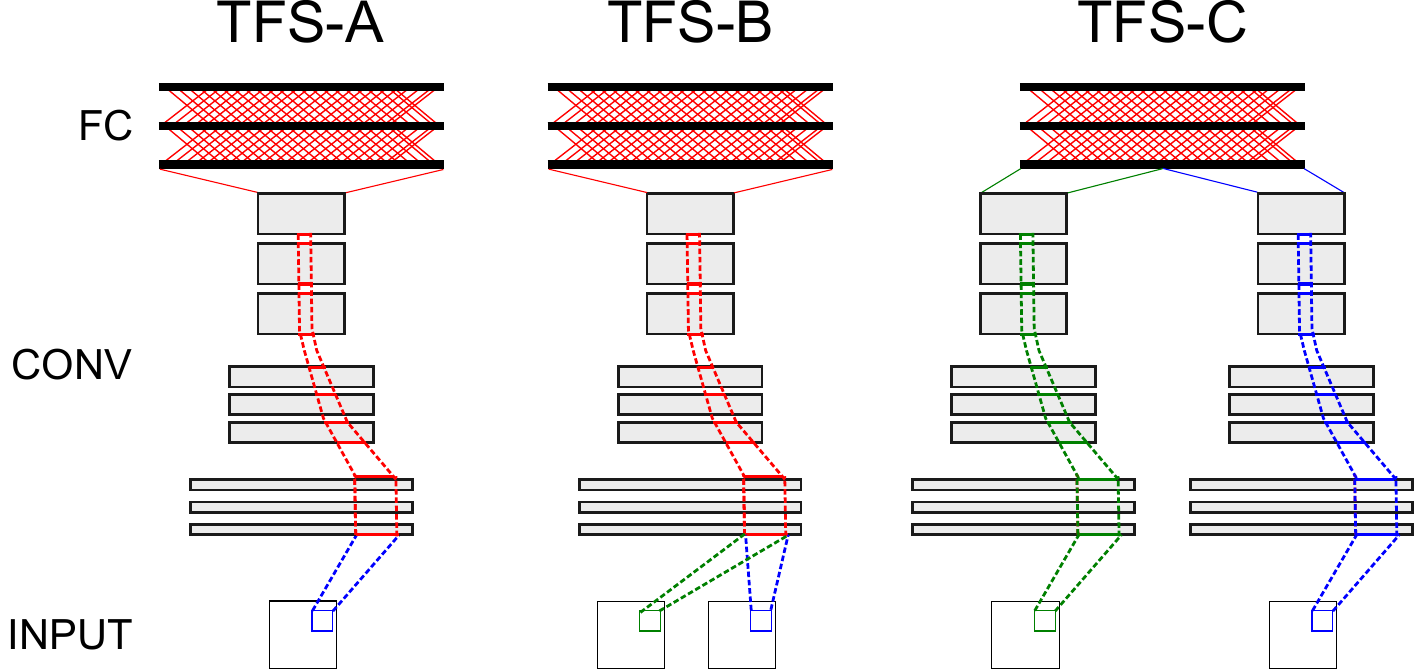}
\caption{Trained-from-scratch (TFS) network configurations evaluated. A. Single channel input images, B. Two channel input images, and C. Two input images feeding into separate convolutional layer streams.}
\label{fig:CNNConfiguration}
\end{figure}

In addition to the computation of the image score $p_{I}$, a heatmap was generated during classification by mapping the normalized mean window score at each location (across all windows overlapping at that location) to pixel values. These visualizations serve two main purposes: i) clarification of classification decision by approximately localizing detected SMTs (or the source of false positive signals), and ii) to serve as a guide to further action by the security officer (e.g. physical inspection).

\subsection{Convolutional Neural Networks}
The main type of CNN evaluated in this contribution were trained-from-scratch (TFS) using the \texttt{MatConvNet} library~\cite{vedaldi2014matconvnet}. Their architecture is based on the very deep networks first described by Simonyan and Zisserman~\cite{Simonyan2014}, where multiple convolutional (CONV) layers with small $3{\times}3$ filters are stacked in-between ``max pooling'' layers and feed forward into three fully-connected (FC) layers. 11-layer (8 CONV + 3 FC)  and 19-layer (16 CONV + 3 FC) variants were explored. For both variants, three configurations were evaluated (Fig.~\ref{fig:CNNConfiguration}): grayscale image input (TFS-A, raw \emph{or} log-transformed intensities); dual channel image input (TFS-B, raw \emph{and} log-transformed intensities); and separate raw and log-transformed inputs to distinct branches of the network (with no weight sharing) whose features are concatenated after the first FC (TFS-C). In all cases, the window score $p_{w,i}$ was given by the output of the softmax layer for the positive class.

Batch normalisation (fixing the mean and variance of input distributions at each layer) was used for network regularisation and to speed up training~\cite{Ioffe2015}. Weight decay and momentum were fixed at ${10^{-4}}$ and 0.9, respectively. Learning rate was decreased from ${10^{-3}}$ to ${10^{-6}}$ over the course of 30 epochs. The mean image computed across the training set was subtracted from each input image. In addition,  images were also randomly flipped (horizontally and/or vertically) at training. 
\begin{table}
\caption{Performance for the detection of SMTs in X-ray cargo images. For clarity, some results were omitted from the table. ``+Log'' denotes that images were log-transformed prior to classification. FPR90 is the false positive rate for a 90\% detection rate.}
\label{tbl:results}
\centering
\begin{tabular}{ l c c c }
\hline
\bf{Method} & \bf{AUC} & \bf{H-measure} & \bf{FPR90} \\ 
\hline 
oBIFs & 0.72 & 0.19 & 0.72\\
oBIFs + Log & 0.59 & 0.04 & 0.88\\
PHOW & 0.72 & 0.18 & 0.75 \\
PHOW + Log & 0.73 & 0.20 & 0.75 \\
\hline
CNN-19-PT-FC1 & 0.67 & 0.17 & 0.86 \\
CNN-19-PT-FC1 + Log & 0.61 & 0.12 & 0.89 \\
CNN-19-PT-FC2 & 0.67 & 0.17 & 0.85 \\
\hline
CNN-11-TFS-A + Log & 0.95 & 0.72 & 0.13 \\
CNN-11-TFS-B & 0.95 & 0.70 & 0.15 \\
\hline
CNN-19-TFS-A & 0.89 & 0.53 & 0.47 \\
CNN-19-TFS-A + Log & 0.96 & 0.75 & 0.09 \\
\textbf{CNN-19-TFS-B} & \textbf{0.97} & \textbf{0.78} & \textbf{0.06} \\
CNN-19-TFS-C & 0.96 & 0.75 & 0.10 \\
\hline\hline
\end{tabular} 
\end{table}

In addition to TFS CNNs, pre-trained (PT) networks were also evaluated. Features were extracted from the FC1 and FC2 layers of a VGG-VD-19\cite{Simonyan2014} model, whose architecture is very similar to the 19-layer TFS CNN, trained on ImageNet (dataset of natural photographic images) and were classified using Random Forest classifiers. Input images were resized to $224{\times}224$ and the grayscale channel was replicated twice in the third dimension to match the expected RGB format. For PT CNNs, the window score $p_{w,i}$ was computed as the fraction of trees voting for the positive class. 

\subsection{Bag-of-Words features}
In addition to CNNs, Bag-of-Words (BoW) features were also evaluated: oriented Basic Image Features (oBIFs) and Pyramid Histograms Of visual Words (PHOW). BIFs are fixed geometric features, classifying each pixel of an image into one of seven categories according to local symmetry~\cite{griffin2009basic}. For this work, we used the extended formulation (oBIFs) where the orientation of rotationally asymmetric features is quantized, resulting in 16 new categories, for a total of 23~\cite{Newell2011}. The oBIF computation was carried out at four scales ($\sigma{=}\{0.7, 1.4, 2.8, 5.6\}$) and two threshold parameters ($\gamma{=}\{0.011, 0.1\}$). These parameters were previously shown to be optimal for detection of cars in cargo containers~\cite{2016arXiv160608078J}. The feature vector for a window was 184-dimensional.
\begin{figure}
\centering
\includegraphics[width=0.94\linewidth]{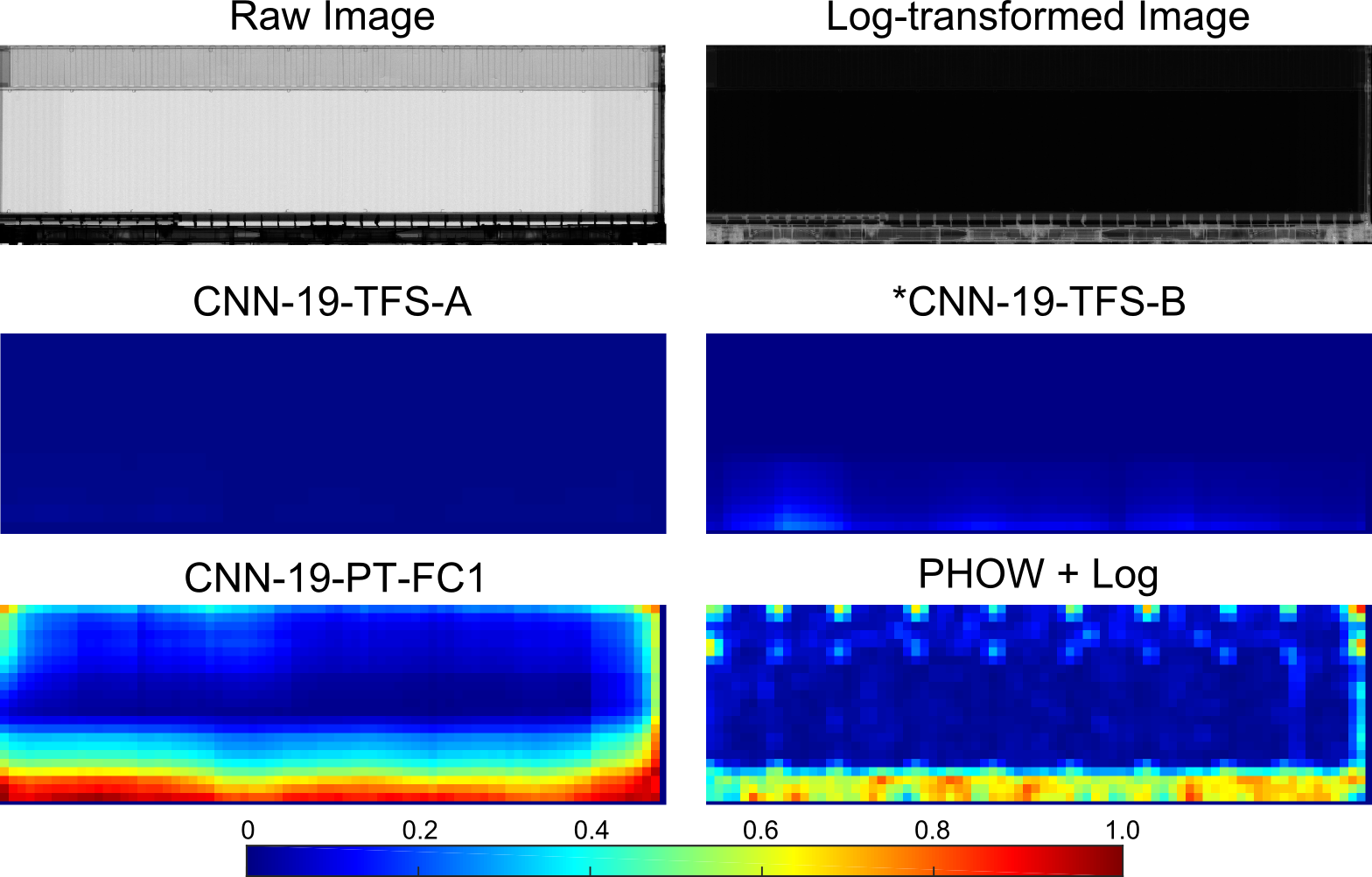}
\caption{SMT detection on an empty container using a selection of the algorithms evaluated. Images have been scaled so that a value of 1.0 (red) corresponds to a false positive detection for a 90\% true positive rate.  The best performing scheme is marked *.}
\label{fig:emptyContainerResponse}
\end{figure}

PHOW were proposed as a multi-scale extension of dense SIFT (Scale-Invariant Feature Transform)~\cite{bosch2006scene,bosch2007image} and are computed as follows: i) computation of dense SIFT for the image considered at four scales (4, 6, 8, and 10 pixel spatial bins); ii) learning of a 300 visual word dictionary by $k$-means clustering of dense SIFT; and iii) computation of a two-level pyramid histogram of visual words ($2{\times}2$ and $4{\times}4$ spatial bins). The resulting feature vector was 6000-dimensional.

Random Forest models were used for classification of images based on oBIFs and PHOW features.
\section{Results}
\label{results}
The SMT detection performance obtained for the different methods evaluated are presented in Table~\ref{tbl:results} and summarized in Table~\ref{tbl:resultsSum}. These results highlight the challenging nature of this classification task. Overall, Bag-of-Words (BoW) methods performed poorly; the best AUC and H-measure was achieved by PHOW on log-transformed inputs while oBIFs had the lowest false positive rate for 90\% detection rate (FPR90) with 72\%. Interestingly, log-transformed inputs slightly increase performance of PHOW but was detrimental to that of oBIFs, potentially due to non-optimal parameter choices.

Pre-trained (PT) CNNs have previously been applied successfully to X-ray imagery and delivered robust baseline performance~\cite{2016arXiv160608078J,akccaytransfer}. However, they generally fared worse than BoW approaches for SMT detection, indicating that generic features that are optimal for natural image classification, and that perform reasonably well for the detection of large objects in X-ray images, are not directly transferable to this task. 
\begin{figure}
\centering
\includegraphics[width=0.94\linewidth]{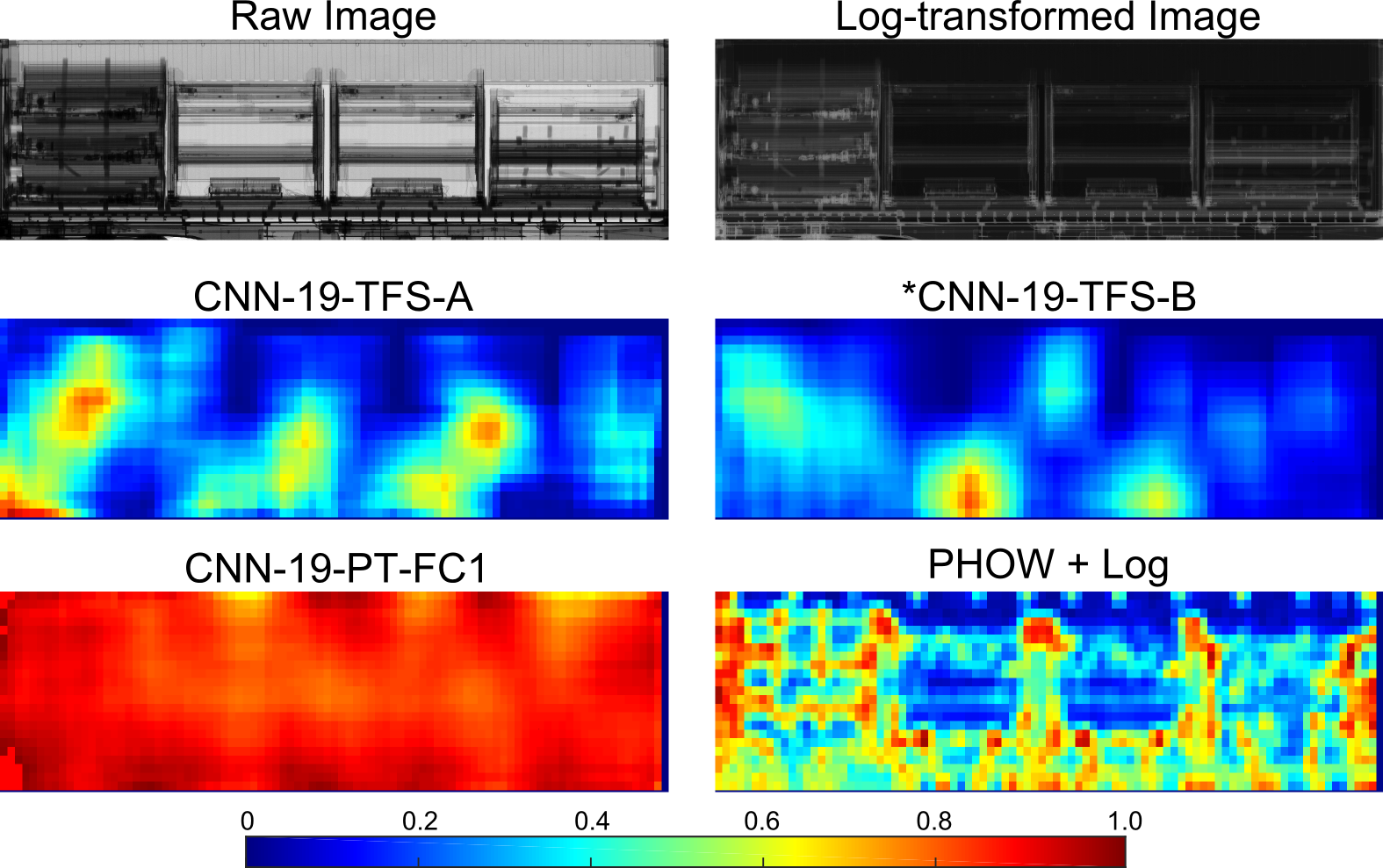}
\caption{SMT detection on a busy container image that does not contain a SMT using a selection of the algorithms evaluated. Images have been scaled so that a value of 1.0 (red) corresponds to a false positive detection for a 90\% true positive rate. The best performing scheme is marked *.}
\label{fig:busyContainerResponse}
\end{figure}

In all cases, trained-from-scratch (TFS) CNNs outperformed both BoW methods and PT CNNs. It was found that log-transforming the image was key in achieving improved performance. For example, log-transforming inputs when using a single channel input (TFS-A) decreased the FPR90 from 47\% down to 9\%. A smaller but still significant improvement was obtained by using inputs with both raw and log-transformed channels (TFS-B), resulting in a further 3\% drop in FPR90 to 6\%. Surprisingly, the network architecture that has two separate streams of convolutional layers for raw and log-transformed input images did not perform better than just using a single log-transformed input (TFS-A + Log). One could expect that encouraging the network to learn channel-specific features would improve classification given the difference in appearance between the two channels. Potentially, this could be explained by the much more complex network over-fitting the training data. The FPR90 was more that doubled when using a shallower network (19-TFS-B versus 11-TFS-B), indicating that the added complexity did not lead to over-fitting in this case.

When processing a benign image of an empty container, the TFS CNNs are the only methods that did not lead to excessive false positive signals (Fig.~\ref{fig:emptyContainerResponse}). Similarly, when given a benign image of a container loaded with industrial equipment and objects, whose appearance closely resemble that of SMTs, PT CNNs and to a lesser degree BoW methods generated very large number of false alarms (Fig.~\ref{fig:busyContainerResponse}). In contrast, only a few image locations had any kind of signal associated with them when using TFS CNNs, and in the case of the dual-channel input variant, no instance was above the threshold to trigger a false alarm.
\begin{figure}[t]
\centering
\includegraphics[width=0.94\linewidth]{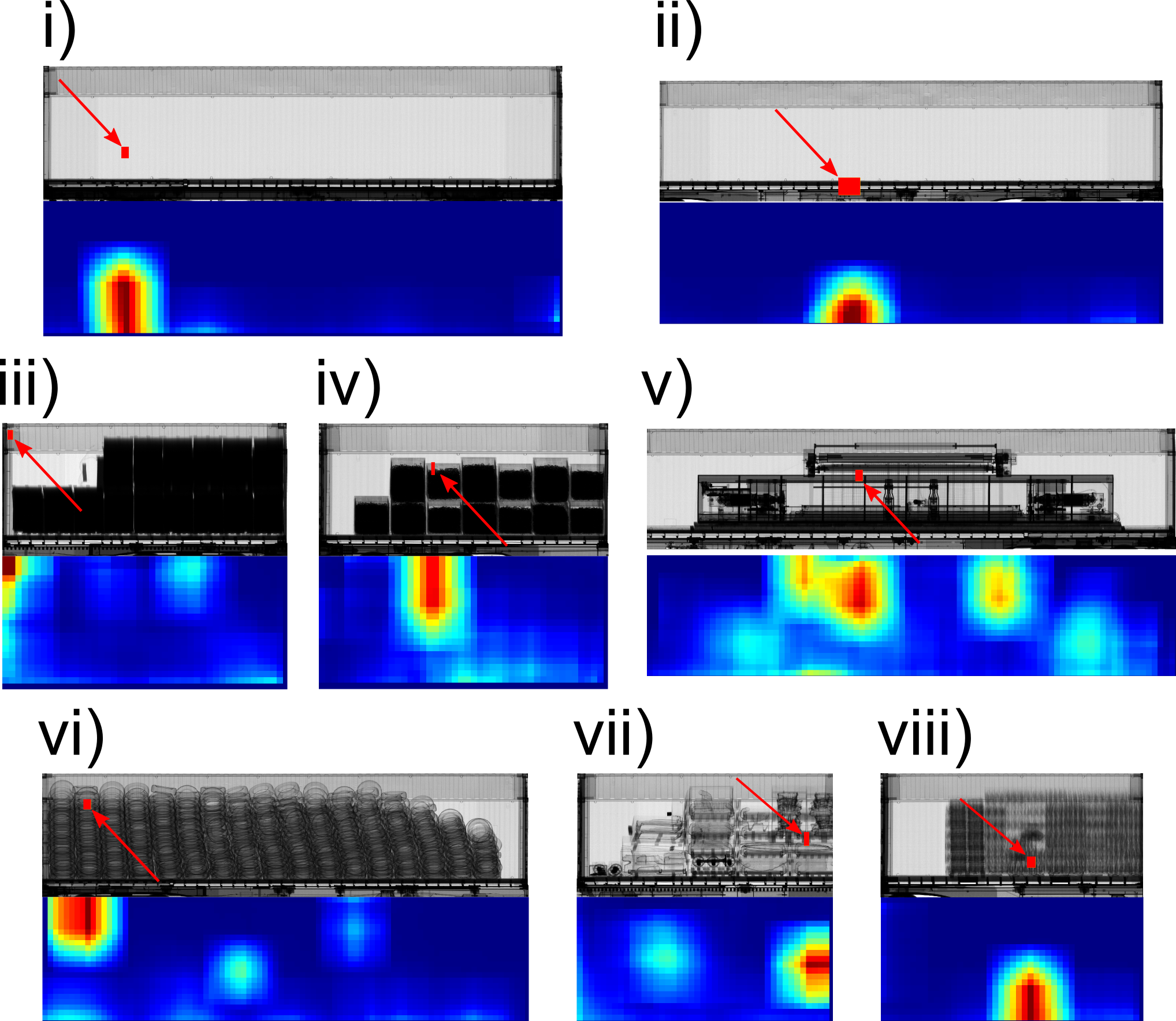}[h]
\caption{SMT detection examples using CNN-19-TFS-B. SMTs are deliberately censored by a red rectangle (the dimension of the rectangles is identical to that of the SMT). i) to iii) shows SMTs concealed in the fabric of the container while iv) and viii) are placed amongst legitimate cargo.}
\label{fig:detectionExamples}
\end{figure}

Examples of successful detections using CNN-19-TFS-B are presented in Figure~\ref{fig:detectionExamples}. In most cases, the signal is well-localized and the classification very specific, especially when projected into empty containers (Fig.~\ref{fig:detectionExamples}.i and ii). The examples where the SMTs are concealed amongst other cargo (Fig.~\ref{fig:detectionExamples}.iii-viii) would be very challenging to detect by visual inspection, especially under time pressure.
\section{Conclusion}
\label{conclusion}
We have proposed a Deep Learning scheme for the detection of ``small metallic threats'' (SMTs) in X-ray cargo images. Using a novel method for the generation of a suitably large and diverse dataset of physically-realistic synthetic images, Convolutional Neural Networks (CNNs) could be trained-from-scratch. We report a 1-in-17 false alarm rate for 90\% detection, which significantly outperforms other methods evaluated, including classification based on pre-trained CNNs and Bag-of-Words features (Table~\ref{tbl:resultsSum}). The processing time using a Titan X GPU was 3.5 second per image in average, which is significantly lower than the time taken by operators to inspect cargo container images.

\begin{table}[h]
\caption{Summary of best performance obtained for each approach (see Table~\ref{tbl:results})}
\label{tbl:resultsSum}
\centering
\begin{tabular}{ l c c c }
\hline
\bf{Method} & \bf{AUC} & \bf{H-measure} & \bf{FPR90} \\ 
\hline 
BoW & 0.72 & 0.19 & 0.72\\
CNN-PT & 0.67 & 0.17 & 0.86\\
CNN-TFS & 0.97 & 0.78 & 0.06 \\
\hline\hline
\end{tabular} 
\end{table}

The scheme described could potentially result in a step change in SMT detection capability. However, further research is required before it is ready to be deployed in the field. Due to the lack of real images containing SMTs concealed amongst legitimate cargo, we have relied on synthetic images for performance evaluation. While all efforts were made to evaluate the system in a way that is meaningful and as representative of real-real world performance as possible (e.g. by using fully disjoint datasets for training and testing, for both threats projected and background patches), it is essential for performance to be evaluated based on real images showing realistic placement of SMTs.

\section*{Acknowledgements}
This work was funded by Rapiscan Systems, and by EPSRC Grant no. EP/G037264/1 as part of UCL's Security Science Doctoral Training Centre.
\small{\bibliography{icdp2009.bib}}
\end{document}